\documentclass[11pt]{article}
\usepackage{adjustbox}

\usepackage[preprint]{acl}

\usepackage{times}
\usepackage{latexsym}
\usepackage{subcaption}
\usepackage{booktabs}
\usepackage{multirow}
\usepackage{float}

\usepackage{amssymb}
\usepackage{amsmath}
\usepackage[dvipsnames]{xcolor}

\usepackage{algorithm}
\usepackage{algpseudocode}
\usepackage{amsmath}

\usepackage[T1]{fontenc}

\usepackage[utf8]{inputenc}

\usepackage{microtype}

\usepackage{inconsolata}

\usepackage{graphicx}

\title{Moment-KV: Momentum-Based Decode-Time KV Cache Compression for Long Generation}

\author{
Soumyadeep Jana\thanks{Equal contribution.}, 
Sagar Nishad\footnotemark[1] \and
Sanasam Ranbir Singh \\
Indian Institute of Technology Guwahati, India \\
\texttt{\{sjana, sagar.nishad, ranbir\}@iitg.ac.in}
}

\begin{document}
\maketitle
\begin{abstract}

Key-Value (KV) cache remains a major bottleneck for deploying Large Language Models (LLMs) in long-generation tasks. Prior work often applies uniform compression across both prefill and decoding caches, but compressing the prefill cache degrades performance by corrupting critical context. While preserving the prefill cache is essential, decoding-phase compression remains underexplored, with existing methods relying on rigid recency windows or instantaneous attention. Our analysis of attention dynamics reveals strong temporal patterns: critical tokens receive sustained attention over long horizons, while local reasoning involves short-lived bursts. Static heuristics fail to capture this behavior, leading to premature eviction of important tokens or retention of stale ones. We propose \textbf{Moment-KV}, a decoding-time KV cache compression method based on momentum-driven temporal attention aggregation. Our method models token importance as a continuously evolving state, where attention is aggregated with decay, capturing both long-term influence and recent relevance. Experiments show that Moment-KV significantly improves generation fidelity in long-generation tasks (\textcolor{ForestGreen}{2.3-3.2 \%} $\uparrow$) while maintaining decoding latency.
\end{abstract}

\section{Introduction}

Large Language Models (LLMs) like LLaMA \cite{Touvron2023LLaMAOA} and Qwen \cite{Bai2023QwenTR}, have shown strong capabilities for text generation. As they scale, their ability to handle longer context windows has improved significantly. This has enabled more effective long-output sequence generation for tasks such as synthesizing large codebases and writing long-form documents.



However, scaling to these extended output lengths introduces major computational challenges, as the Transformer's self-attention grows quadratically ($O(N^2)$) with the number of output tokens, leading to high latency in long-generation tasks. To address this, modern LLMs use a Key-Value (KV) cache that stores previously computed keys and values, avoiding redundant computations during autoregressive generation. While this reduces computation, it creates a memory bottleneck: the KV cache grows linearly ($O(N)$) with sequence length, consuming significant GPU memory in long-generation scenarios. For instance, in a 8B parameter model (e.g., Llama-3.1-8B-Instruct), storing the KV cache in 16-bit precision consumes exactly 128 KB per token. Generating a sequence of 100K tokens requires nearly 13 GB of GPU memory for the KV cache alone, often exceeding the memory footprint of the model weights themselves. To better understand this bottleneck, it is useful to distinguish between two phases of KV caching. In the \textit{prefill phase}, the model processes the input prompt in parallel, constructing a fixed KV cache that encodes the full context. In the \textit{decoding phase}, tokens are generated sequentially, and their KV states are appended at each step, causing the cache to grow unbounded and dominate memory usage.

\begin{figure*}[t!]
    \centering
    \includegraphics[width=0.8\textwidth]{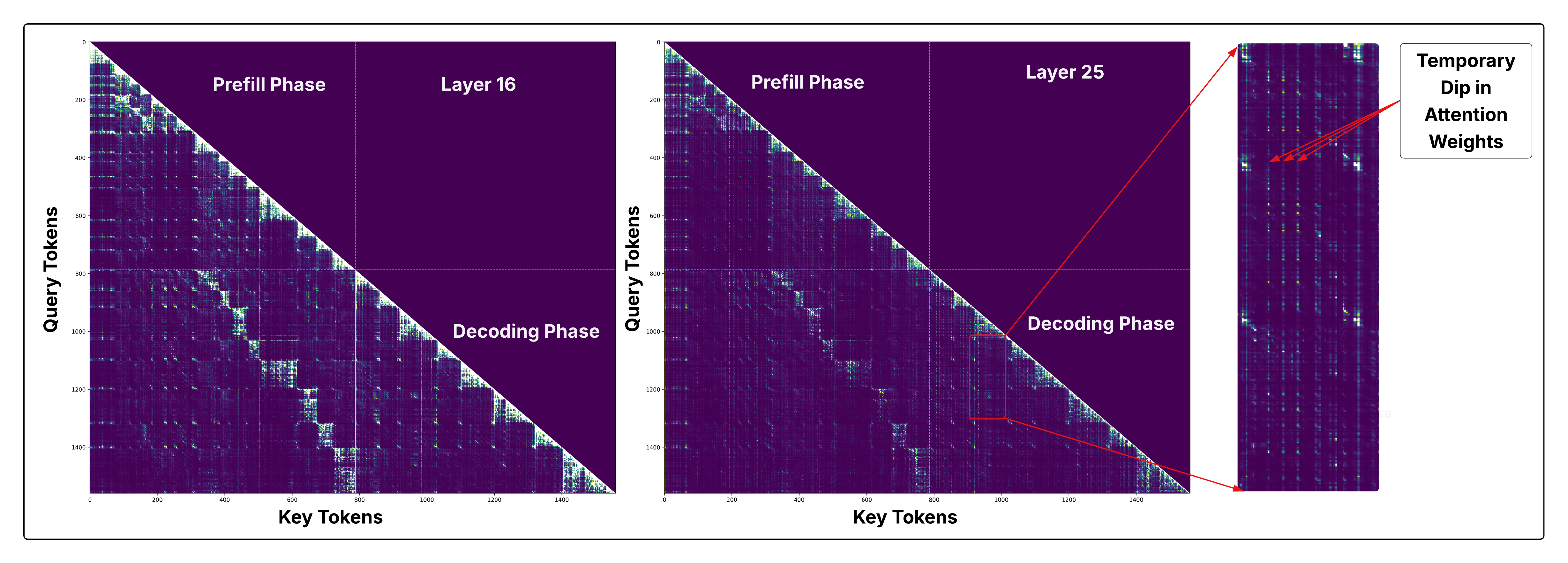}
    \caption{
    Attention Heatmap of LLaMA-3.1-8B-Instruct on GSM8K, showing token attention patterns during prefill and decoding phases for Layers 16 and 25. The zoomed region highlights a temporary dip in attention weights during decoding, where important tokens momentarily receive low attention.}
    \label{fig:attention_pattern}
\end{figure*}

\begin{figure}[t]
    \centering
    \includegraphics[width=0.7\columnwidth]{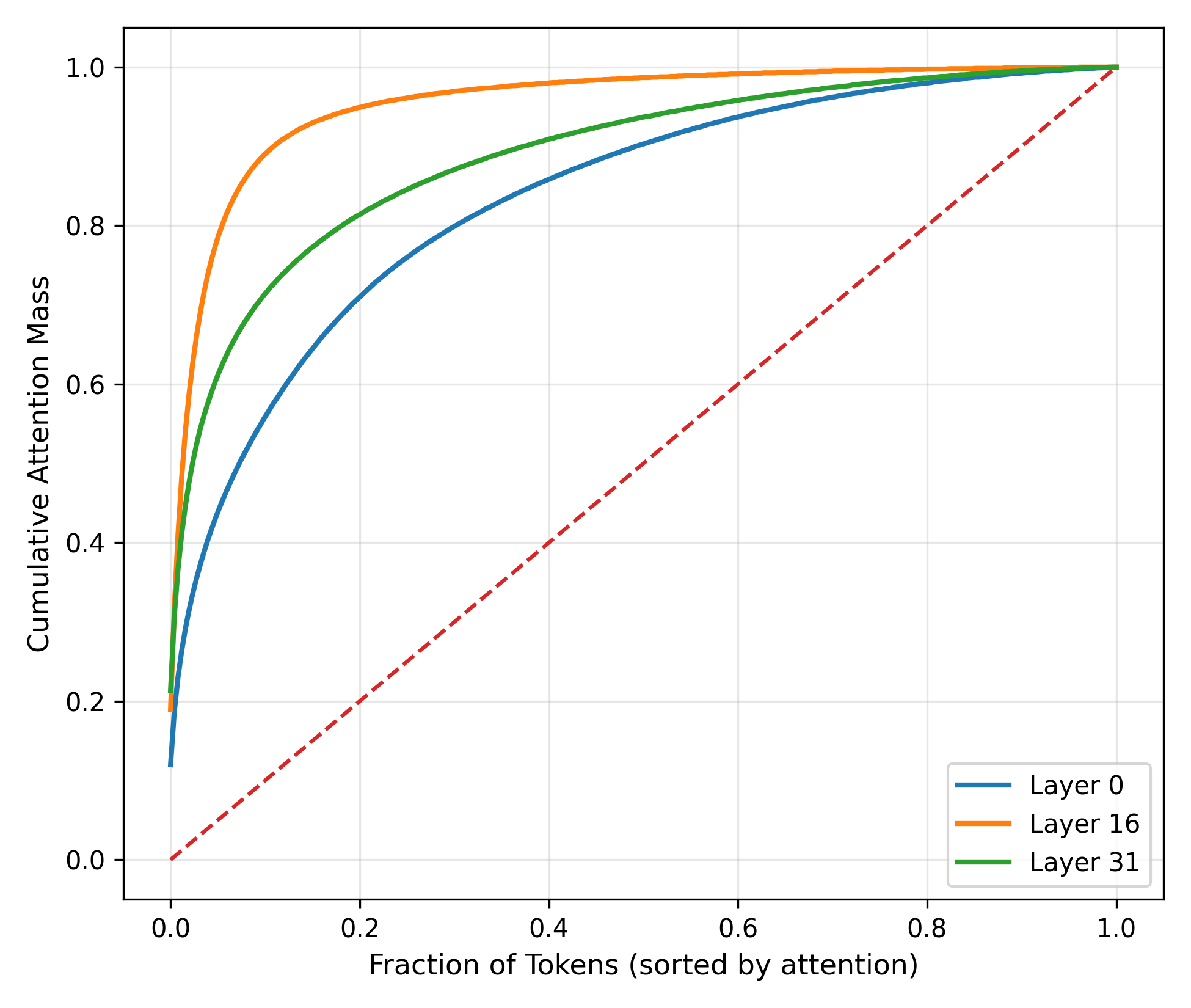}
    \caption{Cumulative attention mass of recently-decoded tokens (window size of 256) for LLaMA-3.1-8B-Instruct on GSM8K.}
    \label{fig:recent_window_cdf}
\end{figure}

Prior work \cite{Li2024SnapKVLK, Cai2024PyramidKVDK, Park2025KeyDiffKS} has largely focused on compressing only the prefill cache or applying unified  method to both prefill and decoding caches \cite{Zhang2023H2OHO, Yang2024PyramidInferPK, Xiao2023EfficientSL}. However, recent studies \cite{Wu2024SCOPEOK, Liu2025FlowKVEM} show that compressing the prefill cache can harm performance, as it preserves critical instructions and contextual semantics. Our experiments reported in Table \ref{tab:prefill+decode} corroborate these findings, highlighting that maintaining full fidelity of the prefill cache is essential for high-quality generation. In contrast, the decoding cache, despite being the primary driver of memory growth in long-generation tasks, remains relatively underexplored, with existing methods \cite{Wu2024SCOPEOK} relying on simple heuristics (e.g., sliding window or recency) that fail to capture the evolving importance of tokens during generation.

To better understand decode-time KV cache behavior, we analyze attention patterns during long-generation in \S \ref{sec:observations} and identify two key limitations of existing methods: \textbf{(L1) Temporal Attention Dynamics:} Attention is inherently temporal. Heavy hitters \footnote{Tokens that receive sustained attention over multiple decoding steps.} influence generation over long horizons but exhibit non-monotonic behavior, with temporary dips in attention before regaining importance. Methods based on instantaneous attention fail to capture this, leading to premature eviction of important tokens. \textbf{(L2) Fixed Recent Window:} Existing methods reserve a fixed portion of the KV cache for recently decoded tokens, assuming recency implies importance. However, attention is often concentrated on only a few recent tokens, causing low-utility tokens to occupy cache space. Together, ignoring temporal dynamics and over-relying on recency leads to suboptimal KV cache utilization.

To address these limitations, we propose \textbf{Moment-KV} (\textbf{Moment}um-Based Decode-Time \textbf{KV} Cache Compression), a decode-time KV cache compression strategy based on momentum-driven temporal attention aggregation. Instead of relying on instantaneous attention, we model token importance as a continuously evolving state, where attention is aggregated with decay over decoding steps. This allows tokens with sustained influence, despite temporary dips, to retain high importance, mitigating premature eviction of critical tokens (L1). At the same time, eviction is performed purely based on aggregated importance, without any recency bias, ensuring that memory is allocated to truly relevant tokens rather than recent or stale ones (L2). Concretely, we freeze the prefill cache and compress only the decoding cache by maintaining importance scores for each token and evicting the lowest-scoring ones when capacity is exceeded. This design adapts to shifting attention patterns, preserving persistent heavy hitters while discarding irrelevant tokens.

We evaluate Moment-KV on two long generation benchmarks, LongGenBench and HelloBench, featuring output lengths ranging from 4K to 16K tokens. Across both moderate (decode budget of 1024 tokens) and extreme (decode budget of 512 tokens) compression settings, Moment-KV consistently outperforms existing baselines, achieving average relative improvements of \textcolor{ForestGreen}{2.32\%} on LongGenBench and \textcolor{ForestGreen}{3.26\%} on HelloBench, while maintaining decoding throughput. Furthermore, Moment-KV is fully compatible with existing prefill-time compression strategies and consistently improves performance by \textcolor{ForestGreen}{7.26\%} when combined with them. Our contributions can be summarized as follows: 

\noindent \textbf{Contributions:} 
\textbf{(1)} We identify key limitations of decode-time KV cache compression, namely temporal attention dynamics and rigid recency bias. 
\textbf{(2)} We propose Moment-KV, a momentum-based framework for decode-time KV cache compression.
\textbf{(3)} We demonstrate consistent gains on long-output benchmarks while maintaining throughput.

\section{Related Works}

 \paragraph{Unified KV Compression:}
Early works primarily adopt unified compression strategies, treating the prefill and decoding stages as a single cache. StreamingLLM \cite{Xiao2023EfficientSL} retains initial attention sinks along with a sliding window of recent tokens to enable long-sequence generation. H2O \cite{Zhang2023H2OHO} tracks cumulative attention to preserve heavy-hitter tokens under a fixed recency buffer, while PyramidInfer \cite{Yang2024PyramidInferPK} exploits layer-wise sparsity to progressively compress the cache. Although effective in reducing memory, these methods fail to distinguish between the distinct roles of prefill and decoding tokens.

\paragraph{Prefill-Time KV Compression:}
Subsequent works focus on compressing the prefill cache during prompt processing. SnapKV \cite{Li2024SnapKVLK} identifies salient KV positions via attention-based voting, while PyramidKV \cite{Cai2024PyramidKVDK} introduces layer-wise capacity allocation. More recent approaches, such as KeyDiff \cite{Park2025KeyDiffKS} and KNorm \cite{devoto-etal-2024-simple}, attempt to further refine prefill compression by selectively pruning tokens based on cosine similarity or key-norm statistics to retain only the most informative representations. Any loss of critical instruction tokens in this phase can significantly degrade reasoning and instruction-following performance.

\paragraph{Decode-Time KV Compression:}
Recent methods such as SCOPE \cite{Wu2024SCOPEOK} decouple the two phases and focus on decoding-time compression. Despite this shift, existing approaches rely on rigid recency heuristics or instantaneous attention signals, failing to capture the temporal dynamics of token importance. Our work addresses this gap by introducing a momentum-based temporal importance formulation, enabling adaptive cache utilization during decoding.

\section{Observations}
\label{sec:observations}
\subsection{Temporal Nature of Attention}

We analyze attention patterns across different layers of LLaMA-3.1 \cite{Dubey2024TheL3} during long-sequence generation. As shown in Figure \ref{fig:attention_pattern}, certain tokens receive sustained attention over many decoding steps. We refer to these as \textit{heavy hitters}, as they persistently influence future tokens over extended horizons, appearing as continuous vertical bands in the attention maps. However, their attention is not strictly monotonic. As highlighted in the zoomed-in region, even important tokens can exhibit temporary dips during local reasoning, before becoming highly relevant again. This temporal behavior reveals a key limitation of existing decoding cache compression methods based on instantaneous attention. By relying on single-step signals, such approaches risk evicting important tokens during brief periods of inactivity, despite their long-term relevance.

\subsection{Curse of Fixed Recent Window}

To better understand how attention is distributed across recently decoded tokens typically retained by existing KV cache compression methods, we analyze the attention mass within a fixed-size recency window (i.e., the last $256$ decoded tokens) in LLaMA-3.1 \cite{Dubey2024TheL3} during long-sequence generation (Figure~\ref{fig:recent_window_cdf}). We observe that attention is highly concentrated: a small fraction of tokens accounts for the majority of attention, as indicated by the steep rise of the curves relative to the uniform baseline. This pattern holds consistently across early, middle, and late layers, with the strongest concentration in middle layers. These results reveal a key inefficiency of fixed KV slots for recent-decoded tokens. Out of all cached tokens, only a small subset meaningfully contributes to generation, while most receive negligible attention. This results in wasted cache capacity for low-utility tokens solely because of their position.

\section{ Methodology }

\subsection{KV Cache Preliminaries}


\noindent \textbf{Prefill Phase:} Given an input prompt $P = \{p_1, p_2, \dots, p_M\}$ of $M$ tokens with hidden dimension $d$, the model processes the entire prompt in a single parallel forward pass. The key and value tensors for the prefill context are computed as:
$ K_P = P W_K, \quad V_P = P W_V$, where $W_K, W_V \in \mathbb{R}^{d \times d}$ are the learned key and value projection matrices. The resulting pair $(K_P, V_P)$ constitutes the prefill KV cache and is written to the cache pool $\Phi^p$. 
The prefill cache encodes the complete contextual representation of the user's prompt.

\noindent\textbf{Decoding Phase:} During the decoding phase, the model generates output tokens autoregressively. At each step $t \in \{1, \dots, T\}$, a new token $x_t$ is processed and its key-value pair is computed as $ K_t = x_t W_K, \quad V_t = x_t W_V$. The pair $(K_t, V_t)$ is appended to the decoding KV cache $\Phi^d$, so the total cache pool at step $t$ is:
\begin{equation}
    \Phi_t = \Phi^p \cup \Phi^d_t, \quad \Phi^d_t = \{ (K_s, V_s) : s \in \{1, \dots, t\} \}
\end{equation}
The attention output for the current query $q_t$ over the full cache is then:



\begin{equation}
\mathrm{Att}_t =
\mathrm{softmax}\left(
\frac{q_t \left(K_{\Phi_t}\right)^\top}{\sqrt{d}}
\right)
V_{\Phi_t}
\end{equation}

\noindent where $K_{\Phi_t}$ and $V_{\Phi_t}$ represents the cached keys and values at decoding timestep $t$. Because $\Phi^d$ grows by one entry at every step, the total cache size at step $T$ is $M + T$, growing linearly with the output length. In long-output scenarios, this unbounded growth rapidly exhausts available GPU memory. 


\subsection{ Moment-KV }
Motivated by the observations in §3, we propose a decoding cache management strategy, presented in Algorithm \ref{alg:momentum_kv}, that preserves the prefill cache while enforcing a strict memory budget on the dynamically growing decoding cache. Rather than evaluating a token importance from a single decode step based on its instantaneous attention , we estimate token's importance through momentum-based attention aggregation across multiple decoding steps. This allows the framework to distinguish tokens that are momentarily unattended but historically critical from tokens that are genuinely obsolete, a distinction that rigid sliding-window or greedy-selection heuristics cannot make. \\

\begin{algorithm}[t]
\caption{Decoding with Moment-KV}
\label{alg:momentum_kv}
\small
\begin{algorithmic}[1]

\Require Current token $x_t$, prefill cache $\Phi_p$, decoding cache $\Phi_d$, importance vector $I$, decoding budget $B_d$, momentum factor $\alpha$

\Ensure Updated cache pool $\Phi_t$

\State $(q_t, K_t, V_t) \gets \textsc{ComputeQKV}(x_t)$

\State $\Phi_d \gets \Phi_d \cup \{(K_t, V_t)\}$

\State $I \gets I \cup \{0\}$

\State $\Phi_t \gets \Phi_p \cup \Phi_d$

\State
$
A_t
\gets
\mathrm{Softmax}
\left(
\frac{q_t K_{\Phi_t}^{\top}}{\sqrt{d}}
\right)
$

\State
$
\bar{A}_t
\gets
\mathrm{Mean}(A_t,\ \mathrm{dim=heads})
$

\State
$
a_{\mathrm{dec}}
\gets
\bar{A}_t[\,|\Phi_p|:\,]
$

\For{$i \gets 1$ to $|\Phi_d|$}

    \State
    $
    I_i(t)
    \gets
    \alpha I_i(t-1) + \bar{a}_i(t)
    $

\EndFor

\If{$|\Phi_d| > B_d$}

    \State
    $
    \mathrm{overflow}
    \gets
    |\Phi_d| - B_d
    $

    \State
    $
    E_t
    \gets
    \operatorname*{arg\,min}_{\mathrm{overflow}}(I)
    $

    \State
    $
    \Phi_d
    \gets
    \Phi_d \setminus E_t
    $

    \State
    $
    I
    \gets
    I \setminus E_t
    $

\EndIf

\State \Return $\Phi_p \cup \Phi_d$

\end{algorithmic}
\end{algorithm}

\noindent \textbf{Cache Architecture}

\noindent We partition the total cache pool into two independent sub-pools:

\paragraph{Prefill pool $\Phi^p$ (frozen):} The complete KV cache produced during the prefill phase is retained in full and never evicted during decoding. Its size $|\Phi^p| = M$ is constant throughout generation. Preserving $\Phi^p$ intact ensures that the model retains full access to the task instruction and contextual grounding, which is essential for reasoning tasks.

\paragraph{Decoding pool $\Phi^d$ (compressed):} All KV pairs generated during the decoding phase are managed in $\Phi^d$, which is subject to a fixed capacity constraint $B_d$. When $|\Phi^d|$ exceeds $B_d$, tokens with the lowest accumulated importance are evicted. The total cache budget at any step $t$ is therefore bounded by:
    
\begin{equation}
    |\Phi_t| \le M + B_d
\end{equation}

\noindent\textbf{Momentum-Based Attention Aggregation}

\noindent At each decoding step $t$, after the new token $x_t$ is appended to the cache, attention is computed over the complete pool $\Phi_t$ . We extract the attention weights restricted to the decoding sub-cache averaged across all $H$ attention heads:
\begin{equation}
    \bar{a}_i(t) = \frac{1}{H} \sum_{h=1}^{H} a_t^h[i], \quad i \in \Phi^d_t
\end{equation}
where $a_t^h[i]$ denotes the attention weight that a token $i$ in the decoding cache recieves at head $h$. Critically, the index $i$ ranges over all decoding tokens currently in the cache, including the freshly appended token $x_t$. 
The per-token importance score is updated as:
\begin{equation}
I_i(t) = \alpha \cdot I_i(t-1) + \bar{a}_i(t)
\end{equation}
where $\alpha \in (0,1)$ is the momentum factor that preserves historical importance while integrating newly observed attention. This momentum-based aggregation allows tokens with sustained attention to build and maintain high importance, while inactive tokens decay toward zero. The parameter $\alpha$ thus controls the memory horizon, with larger values preserving long-term contributions and smaller values emphasizing recent attention. \\

\noindent\textbf{Handling Newly Generated Tokens}

\noindent When a new token $x_t$ is appended to the cache, its importance is initialized to zero: $I_{x_t}(t-1) = 0$. However, eviction is performed \textit{after} the importance update in Eq. (7), not before. By the time capacity is checked, the new token has already received its first importance contribution:
\begin{equation}
    I_{x_t}(t) = \alpha \cdot 0 + \bar{a}_{x_t}(t) = \bar{a}_{x_t}(t) > 0
\end{equation}
The value $\bar{a}_{x_t}(t)$ is the self-attention weight the current query assigns to itself, a strictly positive quantity guaranteed by the softmax operation. In practice, the current query token typically attends strongly to itself, so this initial importance is non-trivial. This design choice prevents trivially evicting freshly generated tokens before they have had any opportunity to demonstrate their relevance, while still permitting eviction of very recent tokens if their self-attention is lower than the accumulated importance of older tokens.

\noindent\textbf{Capacity Enforcement and Eviction}

\noindent After updating importance scores at step $t$, we check whether the decoding cache exceeds its budget:
\begin{equation}
    \text{overflow} = \max(0, |\Phi^d_t| - B_d)
\end{equation}
If $\text{overflow} > 0$, we identify the overflow-many tokens with the lowest importance scores and evict them permanently from $\Phi^d$:
\begin{equation}
    E_t = \arg\min_{\substack{S \subseteq \Phi^d_t \\ |S|=\text{overflow}}} \sum_{i \in S} I_i(t)
\end{equation}
\begin{equation}
    \Phi^d_{t+1} = \Phi^d_t \setminus E_t
\end{equation}
The importance vector is pruned to discard entries of evicted tokens. Eviction is purely score-driven. Any token in the decoding cache, regardless of position or recency, is a candidate for eviction if its accumulated importance is sufficiently low. Conversely, any token with a high historical importance score is retained regardless of when it was generated.

\begin{table*}[!ht]
\centering
\small
\begin{tabular}{l|cccc|cccc}
\toprule
\multirow{2}{*}{Method} & \multicolumn{4}{c|}{LongGenBench-4K} & \multicolumn{4}{c}{LongGenBench-8K} \\
 & GSM8K & MMLU & CSQA & Avg. & GSM8K & MMLU & CSQA & Avg. \\
\midrule
 & \multicolumn{8}{c}{\textbf{LLaMA-3.1-8B-Instruct}} \\
\midrule

Full Cache & 53.18 & 54.27 & 71.54 & 59.63 & 44.51 & 50.92 & 64.81 & 53.41 \\
\midrule
~ & \multicolumn{4}{c|}{Decoding Budget = 1024 Tokens} & \multicolumn{4}{c}{Decoding Budget = 1024 Tokens} \\
\midrule

StreamingLLM & 50.93 & 54.84 & 71.91 & 59.23 & 39.36 & 48.29 & 63.83 & 50.49 \\
H2O & 50.38 & 53.81 & 69.66 & 57.95 & 38.29 & 48.02 & 64.9 & 50.40 \\
PyramidInfer & 46.20 & 53.60 & 69.58 & 56.46 & 34.28 & 45.00 & 57.16 & 45.48 \\
SCOPE & 52.01 & 52.82 &  71.91 & 58.91 & \textbf{44.20} & 49.00  & 66.16 & \textbf{53.12} \\
Moment-KV & \textbf{52.86} & \textbf{55.12} & \textbf{72.83} & \textbf{60.27} & 40.31 & \textbf{50.00} & \textbf{66.33} & 52.21 \\
\midrule
~ & \multicolumn{4}{c|}{Decoding Budget = 512 Tokens} & \multicolumn{4}{c}{Decoding Budget = 512 Tokens} \\
\midrule
StreamingLLM & 44.49 & 55.57 & 71.83 & 57.30 & 33.88 & 48.53 & \textbf{64.25} & 48.89 \\
H2O & 38.75 & 51.42 & 64.58 & 51.58 & 25.07& 44.38 & 47.16 & 38.87\\
PyramidInfer & 37.10 & 51.80 & 63.5 & 50.80 & 24.68 & 42.72 & 52.0 & 39.8 \\
SCOPE & 48.68 & 54.78 & 72.41 & 58.62 & 27.85 & 47.39 & 63.25 & 46.16 \\
Moment-KV & \textbf{50.38} & \textbf{55.63} & \textbf{73.66} & \textbf{59.89} & \textbf{35.63} & \textbf{49.92} & 63.33 & \textbf{49.63} \\

\midrule
& \multicolumn{8}{c}{\textbf{Mistral-7B-Instruct-v0.3}} \\
\midrule

Full Cache & 11.94 & 28.17 & 64.21 & 34.77 & 9.45 & 20.28 & 51.63 & 27.12 \\
\midrule
~ & \multicolumn{4}{c|}{Decoding Budget = 1024 Tokens} & \multicolumn{4}{c}{Decoding Budget = 1024 Tokens} \\
\midrule
StreamingLLM & 11.93 & 27.06 & 64.33 & 34.44 & 6.58 & \textbf{21.08} & 46.5 & 24.72 \\
H2O & 11.24 & 27.81 & 63.91 & 34.32 & 6.03 & 20.22 & 48.41 & 24.89 \\
PyramidInfer & 11.16 & 28.24 & 64.0 & 34.47 & 7.22 & 19.51 & 42.91 & 23.21 \\
SCOPE & 11.39 & 25.09 & 64.75 & 33.74 & 5.79 & 18.00 & 53.25 & 25.68 \\
Moment-KV & \textbf{13.41} & \textbf{28.45} & \textbf{66.04} & \textbf{35.97} & \textbf{8.80} & 19.63 & \textbf{53.33} &  \textbf{27.25}\\

\midrule
~ & \multicolumn{4}{c|}{Decoding Budget = 512 Tokens} & \multicolumn{4}{c}{Decoding Budget = 512 Tokens} \\
\midrule
StreamingLLM & 9.22 & 28.51 & 65.41 & 34.38 & 4.68 &18.64 & \textbf{53.58} & 25.63 \\
H2O & 9.92 & 26.87 & 61.20 & 32.66 & 5.63 & 18.31 & 43.25 & 22.40 \\
PyramidInfer & 9.92 & 26.87 & 61.25 &  32.68 & 5.87 & 17.86 & 39.83 & 21.19 \\
SCOPE & 11.16 & 24.03 & 65.08 & 33.42 & 6.58 & 16.44 & 51.75 & 24.92 \\
Moment-KV & \textbf{12.63} & \textbf{28.57 } & \textbf{65.84} & \textbf{35.68} & \textbf{7.61} & \textbf{19.62} & 50.08 &\textbf{25.77} \\

\bottomrule
\end{tabular}
\caption{Performance comparison with baselines on LongGenBench under different decoding compression settings. Reported metrics denote accuracy(mean of 5 runs). LongGenBench-xk denotes x tokens in the output.}
\end{table*}

\section{Experiments}

\subsection{Benchmarks}
As our method targets decoding-time compression, we focus on benchmarks with long output lengths. We evaluate on \textbf{LongGenBench} \cite{Wu2024SCOPEOK} with output lengths of 4K and 8K tokens, respectively. Each benchmark consists of three question-answering datasets from GSM8K \cite{Cobbe2021TrainingVT}, CSQA \cite{Hendrycks2020MeasuringMM}, and MMLU \cite{Hendrycks2020MeasuringMM}. We further evaluate on the heuristic text generation task from \textbf{HelloBench} \cite{Que2024HelloBenchEL}, with output lengths of 4K and 8K tokens. Additionally, we test on the EnSum task from $\infty$\textbf{Bench} \cite{zhang-etal-2024-bench}, which has an average output length of approximately 1.1K tokens. Additional details are in Appendix \ref{app:datasets}.

\subsection{Baselines}
 We compare \textbf{Moment-KV} against Full Cache and SOTA compression methods, including \textbf{StreamingLLM} \cite{Xiao2023EfficientSL}, \textbf{H2O} \cite{Zhang2023H2OHO} and \textbf{PyramidInfer} \cite{Yang2024PyramidInferPK}, and \textbf{SCOPE} \cite{Wu2024SCOPEOK}. To demonstrate the compatibility of our decoding-based compression method, we further integrate Moment-KV with existing prefill-time compression methods such as \textbf{SnapKV} \cite{Li2024SnapKVLK} and \textbf{PyramidKV} \cite{Yang2024PyramidInferPK}. Baseline details are in Appendix \ref{app:baselines}.

\subsection{Implementation Details}
We conduct experiments using two open-source LLMs, LLaMA-3.1-8B-Instruct and Mistral-7B-Instruct-v0.3. For LongGenBench, we evaluate two output lengths (4K and 8K tokens) and for HelloBench, the output length varies from 2k to 16k tokens. We test our method under two compression settings, retaining 1024 and 512 tokens in $\Phi^d$. The prefill cache $\Phi^p$ is preserved in full and remains uncompressed throughout generation. The momentum factor $\alpha$ is set based on grid-search for different benchmarks. \textbf{To ensure strict parity when comparing against unified compression methods (H2O, StreamingLLM, and PyramidInfer), we constrain their total KV cache capacity (prefill + decode) to the prompt length together with the corresponding decoding budget of either 512 or 1024 tokens}. Hyperparameter details are in the Appendix \ref{app:hyperparameters}.

\section{Results and Discussions}

\subsection{Results on LongGenBench}
Table 1  compares Moment-KV with existing KV cache compression methods on LongGenBench under different decoding budgets and context lengths. Overall, Moment-KV consistently achieves the strongest or highly competitive performance across both LLaMA-3.1-8B-Instruct and Mistral-7B-Instruct-v0.3. Under the 4K setting, Moment-KV achieves the best average performance for both decoding budgets on LLaMA-3.1-8B-Instruct, reaching \textcolor{ForestGreen}{60.27} at a 1024-token budget and \textcolor{ForestGreen}{59.89} at a 512-token budget. In particular, it shows strong gains on reasoning-intensive tasks such as GSM8K and MMLU, indicating that momentum-based importance tracking effectively preserves semantically critical decoding tokens even under aggressive compression. For the more challenging 8K setting, Moment-KV remains highly competitive and consistently outperforms most baselines. At a 512-token budget, it achieves the highest average score among all methods on LLaMA-3.1-8B-Instruct \textcolor{ForestGreen}{49.63}, demonstrating improved robustness under long-context decoding constraints. 

Similar trends are observed on Mistral-7B-Instruct-v0.3. Moment-KV achieves the best overall average performance in nearly all settings, particularly under tighter decoding budgets where token selection becomes more critical. These results suggest that modeling token importance as a temporally evolving momentum state is more effective than relying solely on instantaneous attention statistics.

\subsection{Results on HelloBench}

Table~\ref{tab:cache_budget} presents the performance comparison on the HTG (Heuristic Text Generation) task from HelloBench, which corresponds to the ability of the model to generate original, long-form content. Compared to baselines, the improvements of Moment-KV suggest that it is better at generation-intensive tasks where token importance evolves over time. Notably, Moment-KV with a budget of 1024 achieves the best overall performance at 2K, 4K, and 8K generation lengths, while remaining comparable to Full Cache at 16K.



\subsection{Integration of Moment-KV with Prefill Compression Techniques}

We evaluate the compatibility of Moment-KV with existing prefill compression methods, namely SnapKV and PyramidKV, on the GSM8K task from LongGenBench-4K (Table \ref{tab:prefill+decode}). Using a prefill KV budget of 2048 tokens following \cite{Wu2024SCOPEOK}, Moment-KV consistently outperforms SCOPE across both decode budgets of 1024 and 512 tokens. In particular, Moment-KV maintains stronger performance under aggressive compression (\textcolor{ForestGreen}{25.19} vs \textcolor{red}{22.56} at 512 tokens), demonstrating that momentum-based temporal aggregation provides more robust token retention than instantaneous decode-time heuristics. Overall, Moment-KV acts as an effective plug-and-play decode-time compression strategy that seamlessly complements existing prefill compression methods.

\begin{table}[t]
\centering
\resizebox{\columnwidth}{!}{
\begin{tabular}{l|ccc}
\toprule
\multirow{2}{*}{\textbf{Method}} & \multicolumn{3}{c}{\textbf{Prefill Phase}} \\
 & \textbf{Full Cache} & \textbf{SnapKV} & \textbf{PyramidKV} \\
\midrule

Full Cache & 53.26 & 27.75 & 27.75 \\

\midrule
\multicolumn{4}{c}{Decoding Budget = 1024 Tokens} \\
\midrule

SCOPE & {52.17} & 26.90 & 26.90 \\
Moment-KV & 52.86 & {27.67} & {27.67} \\

\midrule
\multicolumn{4}{c}{Decoding Budget = 512 Tokens} \\
\midrule

SCOPE &  {49.69} & 22.56 & 22.56 \\
Moment-KV & 50.38 & {25.19} & {25.19} \\

\bottomrule
\end{tabular}
}
\caption{Plug-in experiment results of LLaMA-3.1-8B on the GSM8K task from LongGenBench-4K. Prefill budget is 2048 tokens. Reported metric is accuracy.}
\label{tab:prefill+decode}
\end{table}

\begin{table}[ht]
\centering
\renewcommand{\arraystretch}{1.3} 
\resizebox{\columnwidth}{!}{
\begin{tabular}{c c c c c c} 
\toprule
\multirow{2}{*}{\textbf{Method}} & \multirow{2}{*}{\textbf{Budget}} & \multicolumn{4}{c}{\textbf{Generation Length}} \\
\cmidrule(lr){3-6} 
 & & \textbf{2K} & \textbf{4K} & \textbf{8K} & \textbf{16K} \\
\midrule
Full Cache     & $\infty$ & 0.184 & 0.182 & 0.437 & 0.204 \\
\midrule
\multirow{2}{*}{SCOPE} & 512  & 0.186 & 0.181 & 0.431 & 0.188 \\
                               & 1024 & 0.191 & 0.184 & 0.421 & 0.205 \\
\midrule
\multirow{2}{*}{Moment-KV}          & 512  & 0.191 & 0.184 & 0.437 & 0.203 \\
                               & 1024 &  0.201 & 0.192       & 0.445       & 0.199 \\
\bottomrule
\end{tabular}
}
\caption{Performance Comparison of of LLaMA-3.1-8B on HTG task of HelloBench. Budget of 512/1024 indicates the decode-time KV cache size. Reported metric is ROUGE-L.}
\label{tab:cache_budget}
\end{table}

\section{Analysis}

\subsection{Mitigating the Curse of Fixed Recent Window} Figure~\ref{fig:3g_combined}(a) compares the attention distribution over the recent decoding window for SCOPE (we only show for SCOPE as it manages decode KV cache separately) and Moment-KV. SCOPE concentrates most of its attention on only the most recent tokens, while a large portion of the reserved recent window receives very little attention, indicating inefficient utilization of the decoding cache budget. In contrast, Moment-KV maintains a more distributed attention pattern across the window, suggesting that the cache budget is allocated more effectively to tokens that remain useful during generation. These results show that momentum-based temporal aggregation mitigates the limitations of rigid recency-based cache management and improves long-horizon context preservation.

\subsection{Enhanced Long-Horizon Stability} 
\begin{figure*}[t]
    \centering
    \includegraphics[width=\textwidth]{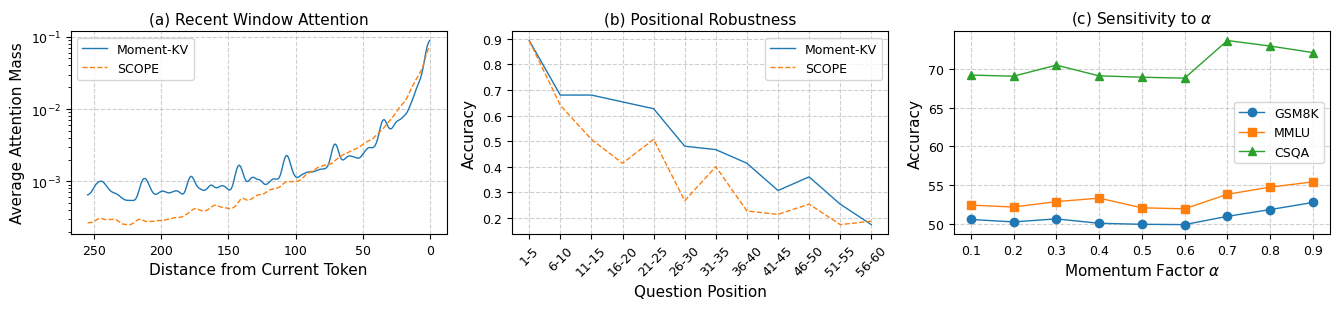}
    \caption{Analysis of Moment-KV across cache utilization, positional robustness, and momentum factor sensitivity.}
    \label{fig:3g_combined}
\end{figure*}

Figure~\ref{fig:3g_combined}(b) shows the accuracy across sequential reasoning steps (Question Index) comparison of our approach vs SCOPE (we only show for SCOPE as it manages decode KV cache separately), for long answer generation in GSM8K dataset. As the context lengthens and the task becomes more complex, both models naturally experience a decline in accuracy. However SCOPE suffers a rapid and severe degradation. By comparison, Moment-KV exhibits significantly higher robustness, maintaining a consistently higher accuracy curve relative to the baseline at advanced question indices. This demonstrates that Moment KV effectively preserves the critical long-term context necessary to support complex, multi-turn generation over extended horizons.

\subsection{Sensitivity of Momentum Factor $\alpha$} 

The momentum factor $\alpha$ controls the temporal memory horizon of Moment-KV by balancing historical importance against newly observed attention. Larger values of $\alpha$ retain historical attention contributions for longer durations, while smaller values emphasize more recent attention patterns and cause token importance to decay more aggressively. Figure~\ref{fig:3g_combined}(c) shows the effect of varying $\alpha$ across the GSM8K, MMLU, and CSQA subsets of LongGenBench-4K for LLaMA-3.1-8B-Instruct under a decoding budget of 512 tokens. Across all three datasets, we observe consistently stronger performance for relatively larger values of $\alpha$, suggesting that long-form generation benefits from preserving temporally persistent \textit{heavy hitters} tokens rather than relying primarily on short-term attention dynamics. At the same time, performance remains relatively stable across a broad range of $\alpha$ values, demonstrating that Moment-KV is not overly sensitive to precise hyperparameter selection.



\subsection{Throughput Analysis}

\begin{figure}[t]
    \centering
    \includegraphics[width=0.7\columnwidth]{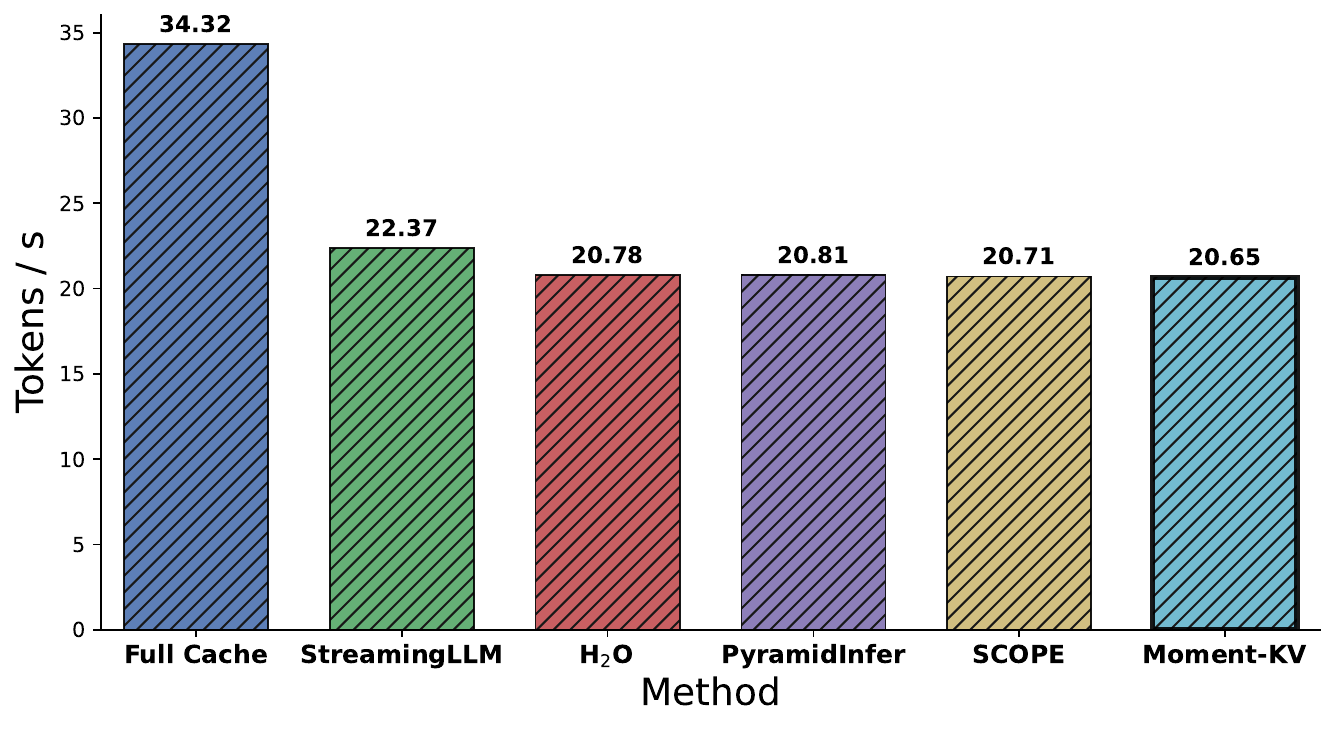}
    \caption{Throughput comparison across KV cache compression methods. Moment-KV achieves competitive performance under constrained memory budgets.}
    \label{fig:throughput}
\end{figure}

Figure~\ref{fig:throughput} compares the decoding throughput of different methods. Full Cache achieves the highest throughput (34.32 tokens/s), as it avoids any cache management overhead. Among compressed methods, StreamingLLM (22.37 tokens/s) and PyramidInfer (20.81 tokens/s) are slightly faster due to their lightweight heuristics, while H$_2$O  (20.78 tokens/s) and SCOPE (20.71 tokens/s) incur minor overhead, due to cumulative attention score computation. Moment-KV achieves competitive throughput (20.65 tokens/s), closely matching existing methods while introducing minimal overhead despite its momentum-based importance tracking.

\subsection{Generalization of Moment-KV}
\begin{figure}[t]
    \centering
    \includegraphics[width=0.7\columnwidth]{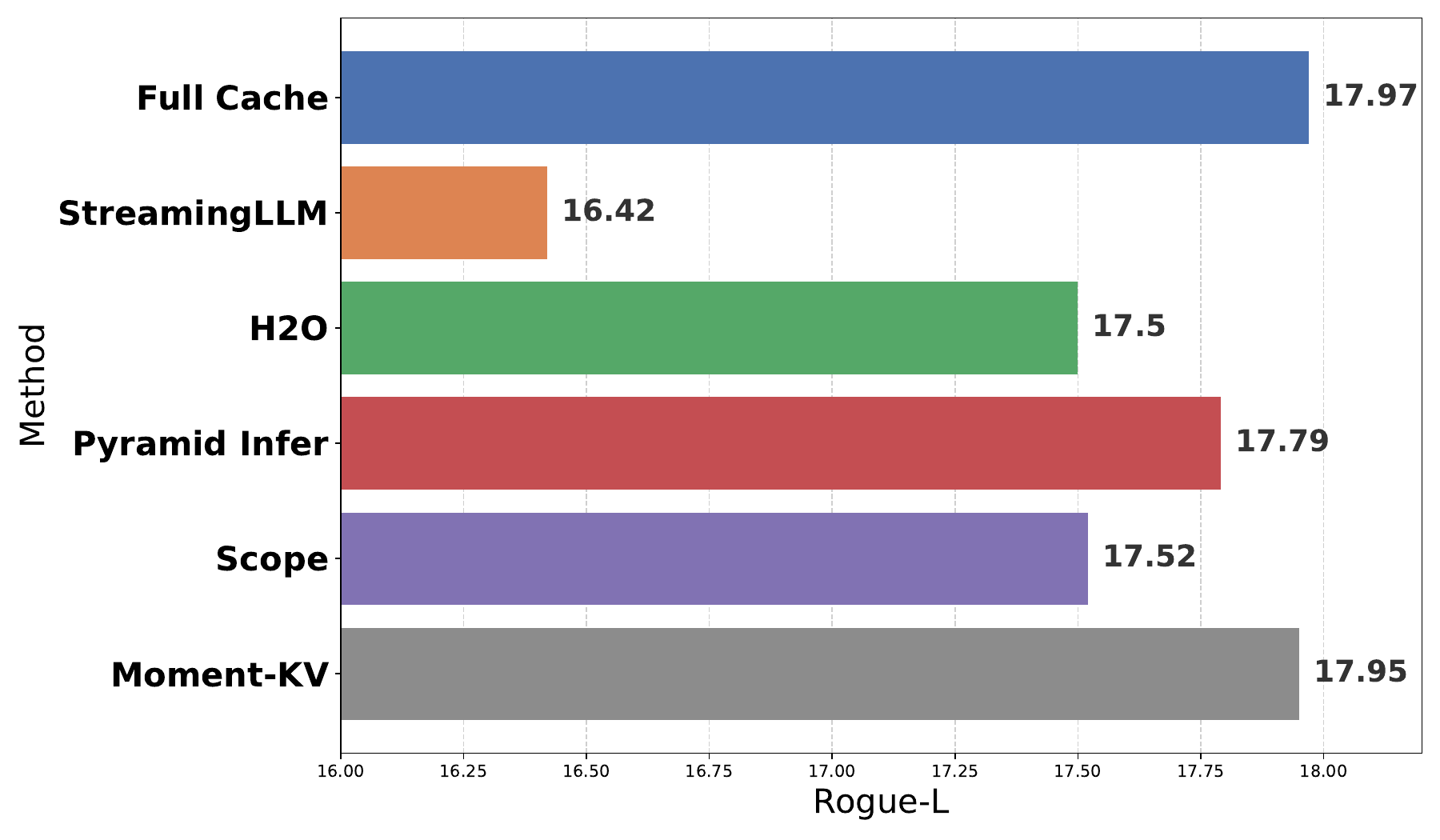}
    \caption{Performance comparison on En-Sum task of $\infty$Bench.}
    \label{fig:infinitebench}
\end{figure}

Results on the En-Sum task in $\infty$Bench (Figure~\ref{fig:infinitebench}) show that Moment-KV achieves the highest performance among all compression methods, nearly matching the full-cache setting. It consistently outperforms SCOPE and other baselines, highlighting the advantage of momentum-based temporal importance over heuristic or instantaneous strategies. These results further confirm that our approach generalizes effectively to long-form generation tasks such as summarization under constrained memory.

\section{Conclusion}
We introduced Moment-KV, one of the first decode-time KV cache compression frameworks for long-sequence generation. We identify two major limitations of existing methods: the temporal instability of attention and the reliance on fixed-size recency windows that can evict important tokens while retaining stale ones. To address this, Moment-KV models token importance as a momentum-driven evolving state by aggregating attention over decoding steps with decay. Extensive experiments on LongGenBench and HelloBench show that Moment-KV consistently outperforms existing baselines under aggressive KV compression, while introducing minimal latency overhead and remaining fully compatible with existing prefill-time compression methods.

\section*{Limitations}
While Moment-KV demonstrates strong performance in preserving long-horizon context during decoding, several directions remain for future work. As one of the first methods targeting decode-time KV cache compression, Moment-KV currently relies on self-attention as a practical proxy for token importance, which may not always perfectly capture true semantic relevance. Additionally, all newly generated tokens are initialized uniformly and subjected to the same momentum aggregation, despite differences in their informational significance. Future work can explore richer importance estimation strategies and adaptive token initialization mechanisms for further improving cache retention decisions.

\bibliography{custom}

\appendix
\section{Dataset Description}
\label{app:datasets}
Throughout our experiments on long-context generation, we evaluate our method on three diverse benchmarks: \textbf{LongGenBench}, \textbf{InfiniteBench}, and \textbf{HelloBench}. These benchmarks collectively assess a broad range of long-context reasoning and long-form generation capabilities. The characteristics and evaluated sub-tasks of each benchmark are described below.
\subsection{LongGenBench}
LongGenBench restructures traditional evaluation formats by sequentially concatenating a batch of questions. This requires the model to generate a cohesive, singular long-context response that addresses each question in order. We utilize all three core domains synthesized within this benchmark: GSM8K, MMLU, and CSQA. The specific structural configurations for both the 4K and 8K variants, delineating the number of concatenated questions per query ($K$) alongside their respective evaluation iteration counts ($T$), are comprehensively detailed in Table~\ref{tab:config}:

\paragraph{GSM8K}: Evaluates arithmetic problem-solving skills using grade-school-level math word problems.
\paragraph{MMLU}: Measures world knowledge and reasoning capabilities across 57 diverse categories.
\paragraph{CSQA} : Tests commonsense reasoning capabilities using questions grounded in ConceptNet.


\subsection{InfiniteBench}
InfiniteBench evaluates LLM capabilities on extreme context lengths surpassing 100K tokens. For our evaluation, we focus on the Text Summarization (\textit{En.Sum}) task, utilizing 103 examples from the official repository. This task requires the model to read an entire novel and generate a concise summary. To accommodate the maximum context window limits of the evaluated models, we apply a bidirectional truncation strategy to oversized inputs. If a tokenized prompt exceeds the model's maximum capacity, we extract the first 50\% and the final 50\% of the allowed token budget, dropping the intermediate text. This ensures the model retains the initial system instructions and the final querying prompt while strictly adhering to memory constraints.

\subsection{HelloBench}
HelloBench is a comprehensive benchmark for evaluating open-ended long-text generation based on Bloom's Taxonomy~\citep{Que2024HelloBenchEL}. We focus on the Heuristic Text Generation task, which corresponds to the ``Create'' cognitive level and evaluates a model's ability to generate coherent long-form content from heuristic prompts. Since HelloBench does not provide reference outputs, we use completions generated by Meta's LLaMA-3.1-70B as pseudo-reference texts for evaluation and report ROUGE-L scores.




\section{Prompt Templates}
\label{app:prompt_templates}

\begin{figure}[t]
    \centering
    \includegraphics[width=1\linewidth]{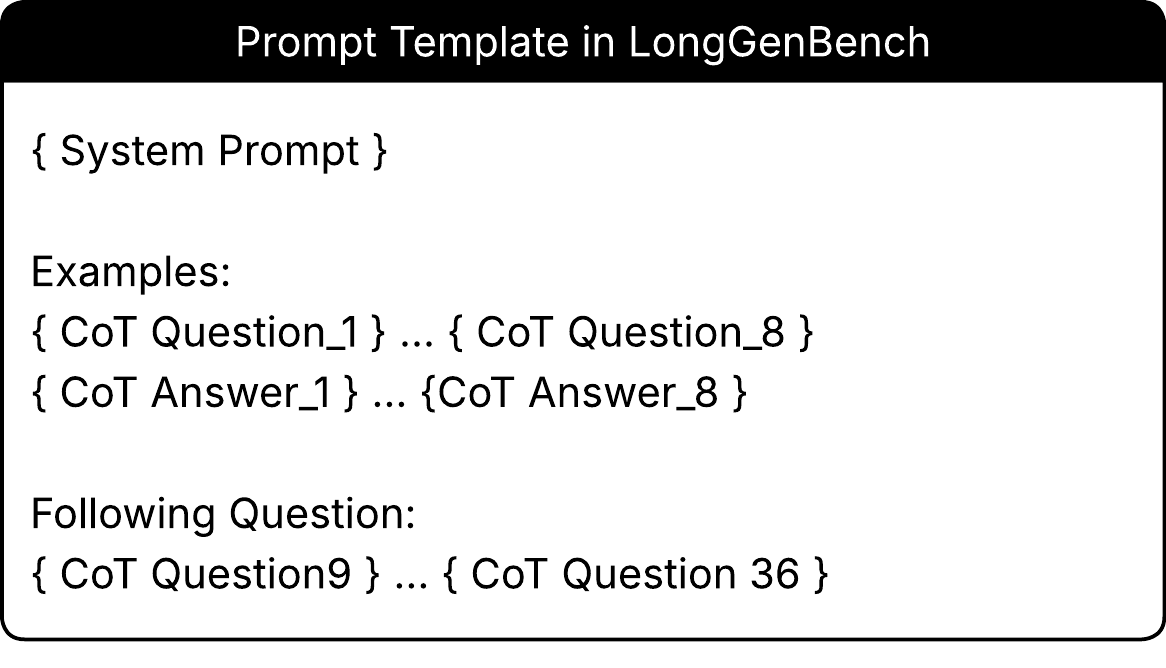}
    \refstepcounter{figure}
    \label{fig:prompt_longgen}
\end{figure}

\begin{figure}[t]
    \centering
    \includegraphics[width=1\linewidth]{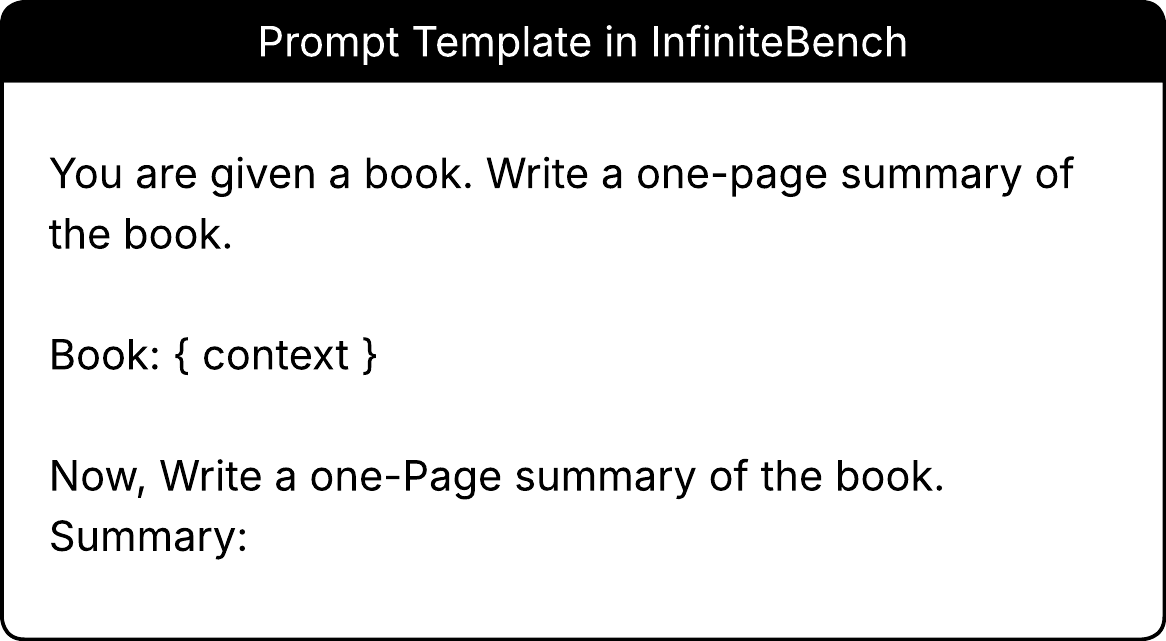}
    \refstepcounter{figure}
    \label{fig:prompt_infinite}
\end{figure}

\begin{figure}[t]
    \centering
    \includegraphics[width=1\linewidth]{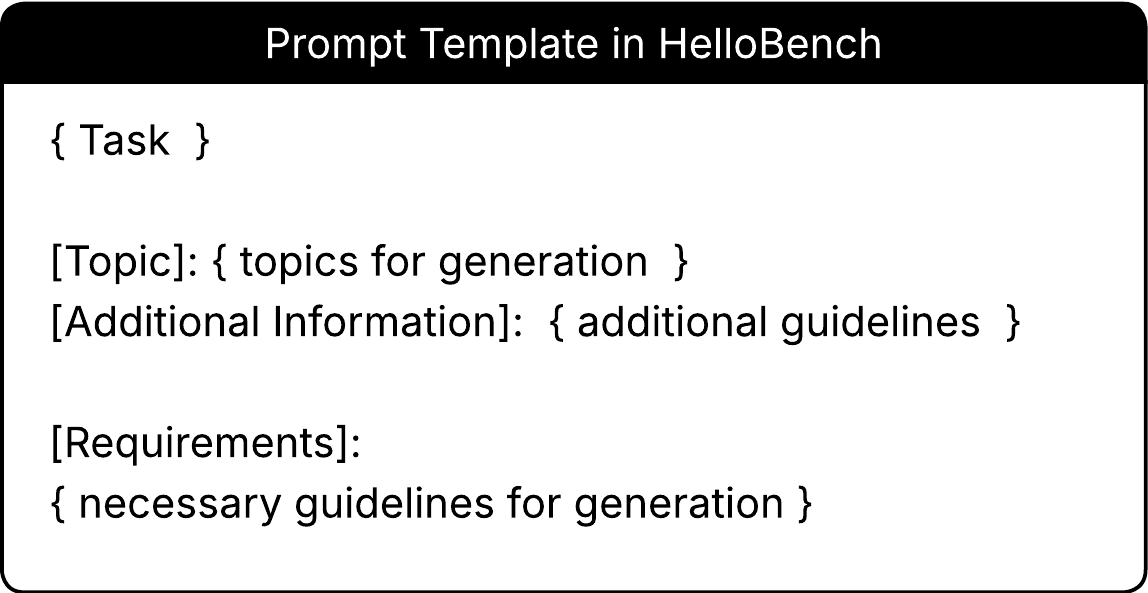}
    \refstepcounter{figure}
    \label{fig:prompt_hello}
\end{figure}

This section outlines the exact prompt templates utilized across all benchmark datasets. Figures \ref{fig:prompt_longgen}, \ref{fig:prompt_infinite}, and \ref{fig:prompt_hello} detail the specific templates used for LongGenBench, InfiniteBench, and HelloBench, respectively. 

\begin{table}[t]
\centering
\resizebox{1\columnwidth}{!}{
\begin{tabular}{l|cc|cc|cc}
\midrule
\multirow{2}{*}{\textbf{Type}} & \multicolumn{2}{c|}{\textbf{GSM8K}} & \multicolumn{2}{c|}{\textbf{MMLU}} & \multicolumn{2}{c}{\textbf{CSQA}} \\
 & $K$ & $T$ & $K$ & $T$ & $K$ & $T$ \\
\hline
LongGenBench-4K & 30 & 129 & 30 & 110 & 40 & 60 \\
LongGenBench-8K & 60 & 42 & 60 & 167 & 80 & 30 \\
\midrule
\end{tabular}
}
\caption{Configuration details for the experiment. The table shows the number of questions in one query ($K$) and the number of iteration times ($T$).}
\label{tab:config}
\end{table}





\begin{table}[t]
\centering
\renewcommand{\arraystretch}{1.2}
\setlength{\tabcolsep}{6pt}

\resizebox{\columnwidth}{!}{%
\begin{tabular}{lcccccc}
\toprule

\multirow{2}{*}{$B_d$}
& \multicolumn{3}{c}{LongGenBench-4K}
& \multicolumn{3}{c}{LongGenBench-8K} \\

\cmidrule(lr){2-4}
\cmidrule(lr){5-7}

& GSM8K & MMLU & CSQA
& GSM8K & MMLU & CSQA \\

\midrule

\multicolumn{7}{c}{\textbf{LLaMA-3.1-8B-Instruct}} \\
\midrule

1024 & 0.98 & 0.98 & 0.55 & 0.97 & 0.97 & 0.80 \\
512  & 0.98 & 0.96 & 0.75 & 0.97 & 0.75 & 0.30 \\

\midrule

\multicolumn{7}{c}{\textbf{Mistral-7B-Instruct-v0.3}} \\
\midrule

1024 & 0.97 & 0.97 & 0.75 & 0.95 & 0.95 & 0.95 \\
512  & 0.90 & 0.95 & 0.83 & 0.95 & 0.95 & 0.97 \\

\bottomrule
\end{tabular}%
}

\caption{
Selected momentum factor $\alpha$ across different LongGenBench experiments.
}

\label{tab:alpha_values_main_table}
\end{table}














\begin{table}[t]
\centering

\begin{tabular}{cccc}
\toprule

\textbf{$B_d$} & \textbf{Full Cache} & \textbf{SnapKV} & \textbf{PyramidKV} \\

\midrule

1024 & 0.98 & 0.97 & 0.97 \\

512  & 0.98 & 0.97 & 0.97 \\

\bottomrule
\end{tabular}

\caption{
Selected momentum factor $\alpha$ for plug-in cache experiments using LLaMA-3.1-8B-Instruct on the GSM8K task from LongGenBench-4K.
}

\label{tab:alpha_values_plugin}
\end{table}
\begin{table}[t]
\centering

\renewcommand{\arraystretch}{1.2}
\setlength{\tabcolsep}{12pt}

\begin{tabular}{ccccc}
\toprule

\multirow{2}{*}{$B_d$}
& \multicolumn{4}{c}{Generation Length} \\

\cmidrule(lr){2-5}

& 2K & 4K & 8K & 16K \\

\midrule

512  & 0.65 & 0.20 & 0.95 & 0.25 \\
1024 & 0.95 & 0.65 & 0.65 & 0.95 \\

\bottomrule
\end{tabular}

\caption{
Selected momentum factor $\alpha$ for HelloBench experiments with LLaMA-3.1-8B-Instruct.
}

\label{tab:alpha_values_hellobench}

\end{table}

\section{Baselines}
\label{app:baselines}

We compare Moment-KV against representative KV cache compression methods spanning three categories: unified KV compression, prefill-time compression, and decoding-time compression.

\paragraph{Unified KV Compression.}
Unified compression methods apply cache management jointly over both the prefill and decoding phases without explicitly distinguishing their functional roles. We evaluate against:
\textbf{StreamingLLM}~\citep{Xiao2023EfficientSL}, which preserves attention sink tokens together with a sliding window of recent tokens;
\textbf{H2O}~\citep{Zhang2023H2OHO}, which retains heavy-hitter tokens based on accumulated attention statistics while reserving a fixed recency window; and
\textbf{PyramidInfer}~\citep{Yang2024PyramidInferPK}, which progressively compresses the KV cache using layer-wise sparsity patterns.
For fair comparison, these methods are evaluated under the same total KV cache budget as Moment-KV, i.e., the prompt length plus the designated decoding budget (512 or 1024 tokens).

\paragraph{Prefill-Time KV Compression.}
Prefill compression methods reduce the KV cache during prompt processing while leaving the decoding process unchanged. We consider:
\textbf{SnapKV}~\citep{Li2024SnapKVLK}, which identifies salient prompt tokens using attention-based voting; and
\textbf{PyramidKV}~\citep{Cai2024PyramidKVDK}, which allocates compression capacity across layers using a pyramidal strategy.
Since Moment-KV is a decoding-only compression framework, these methods are integrated with our decoding strategy to evaluate compatibility and modularity. Specifically, we preserve the compressed prefill cache produced by these methods and apply Moment-KV only during decoding under the same decoding cache budget.

\paragraph{Decoding-Time KV Compression.}
Our primary comparison is with \textbf{SCOPE}~\citep{Wu2024SCOPEOK}, a decoding-time KV cache compression method that preserves the prefill cache while compressing only the decoding cache. SCOPE employs a sliding-window based eviction strategy guided by instantaneous attention and recency heuristics. Since both SCOPE and Moment-KV exclusively target the decoding phase while preserving the prefill cache intact, this comparison isolates the effectiveness of different decoding-time cache management strategies under identical memory budgets.

\paragraph{Full Cache.}
We additionally report results using Full Cache, where no KV compression is applied and all prefill and decoding tokens are retained throughout generation.


\section{Hyperparameter Details}
\label{app:hyperparameters}

In our Moment-KV token importance tracking, we utilize a single primary hyperparameter: the momentum factor \(\alpha\). This parameter controls the decay of historical importance while integrating newly observed attention scores. The importance \(I_i(t)\) of token \(i\) at decoding step \(t\) is updated as follows:
\[
I_i(t) = \alpha I_i(t-1) + a_i(t)
\]
where \(a_i(t)\) represents the attention score of token \(i\) at the current decoding step. Exact hyperparameter settings used across different datasets and models are provided in Table~\ref{tab:alpha_values_main_table}, ~\ref{tab:alpha_values_plugin}, and ~\ref{tab:alpha_values_hellobench}. All hardware evaluations and experiments are conducted using an NVIDIA A100 80GB GPU.

\paragraph{Budget Setup}
To ensure a fair and standardized evaluation framework across all experiments, we establish a strict KV cache budget allocation strategy. Our approach distinctly separates the memory budgets for the prefill and decoding phases.

\paragraph{Prefill Phase:}We do not apply any compression to the KV cache during the initial prefill stage. Therefore, the cache budget allocated for this phase is dynamically set to equal the exact number of tokens present in the input prompt.
\paragraph{Decoding Phase:}During the autoregressive decoding phase, we maintain a strictly bounded decoding cache budget of either 512 or 1024 tokens across all experimental configurations.

Consequently, the total memory footprint utilized by our proposed method at any given step is explicitly limited to the sum of the initial prompt length and the designated decoding budget (i.e., Prompt Length + 512, or Prompt Length + 1024).

\paragraph{Baseline Comparison Strategy} 
To maintain strict parity when evaluating against baseline unified compression algorithms, we constrain their total available KV cache to this exact same global limit. Specifically, the maximum allowable KV cache for any unified compression baseline is strictly restricted so that it never exceeds the combined capacity of the prompt length plus the respective 512 or 1024 decoding token budget.

\clearpage

\begin{figure}[t]
    \centering
    \includegraphics[width=2\linewidth]{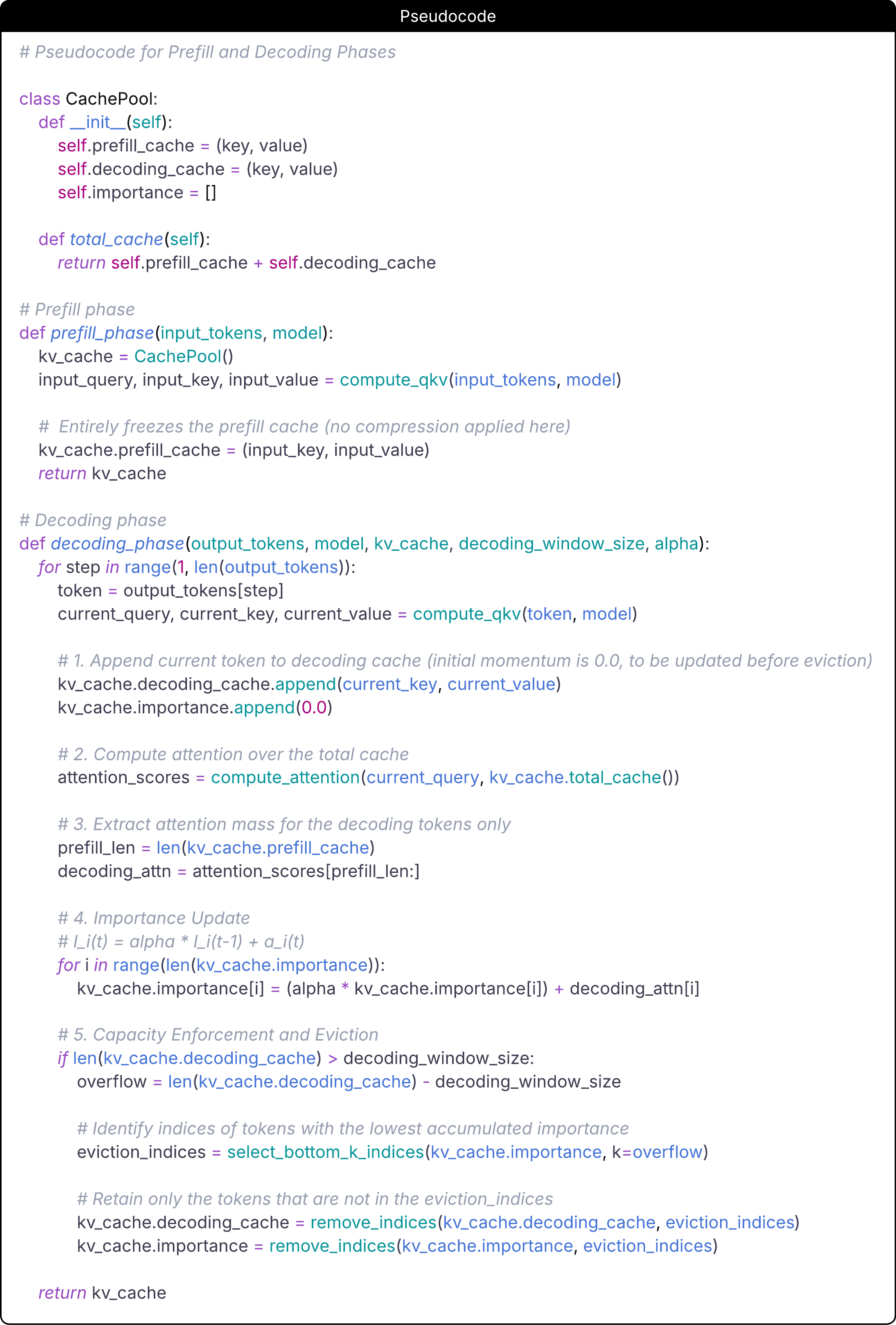}
    \label{fig:pseudocode}
\end{figure}

\clearpage 

\begin{figure}[H]
    \centering
    \includegraphics[width=2\linewidth]{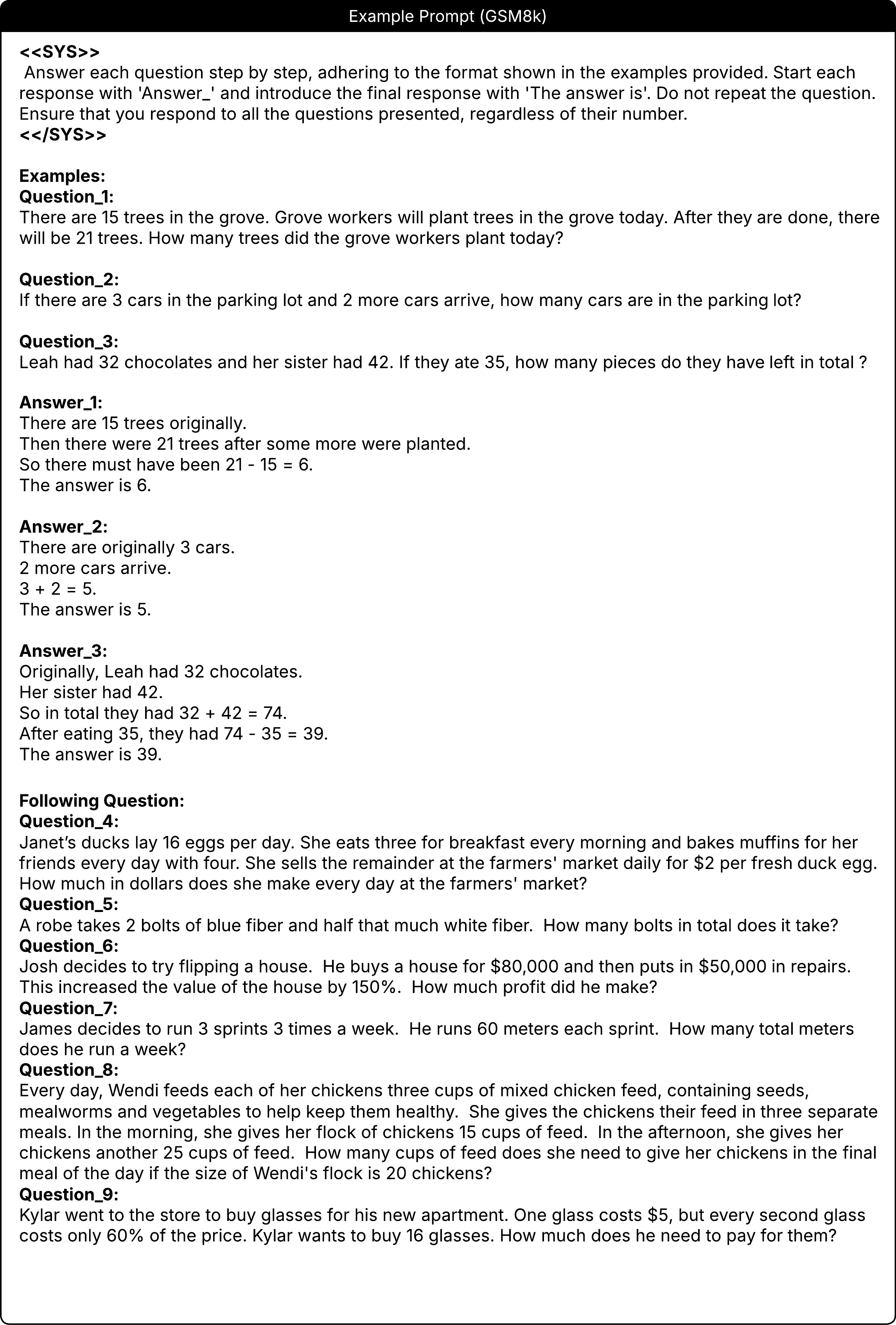}
    \label{fig:gsm8k_prompt}
\end{figure}

\clearpage 

\begin{figure}
    \centering
    \includegraphics[width=2\linewidth]{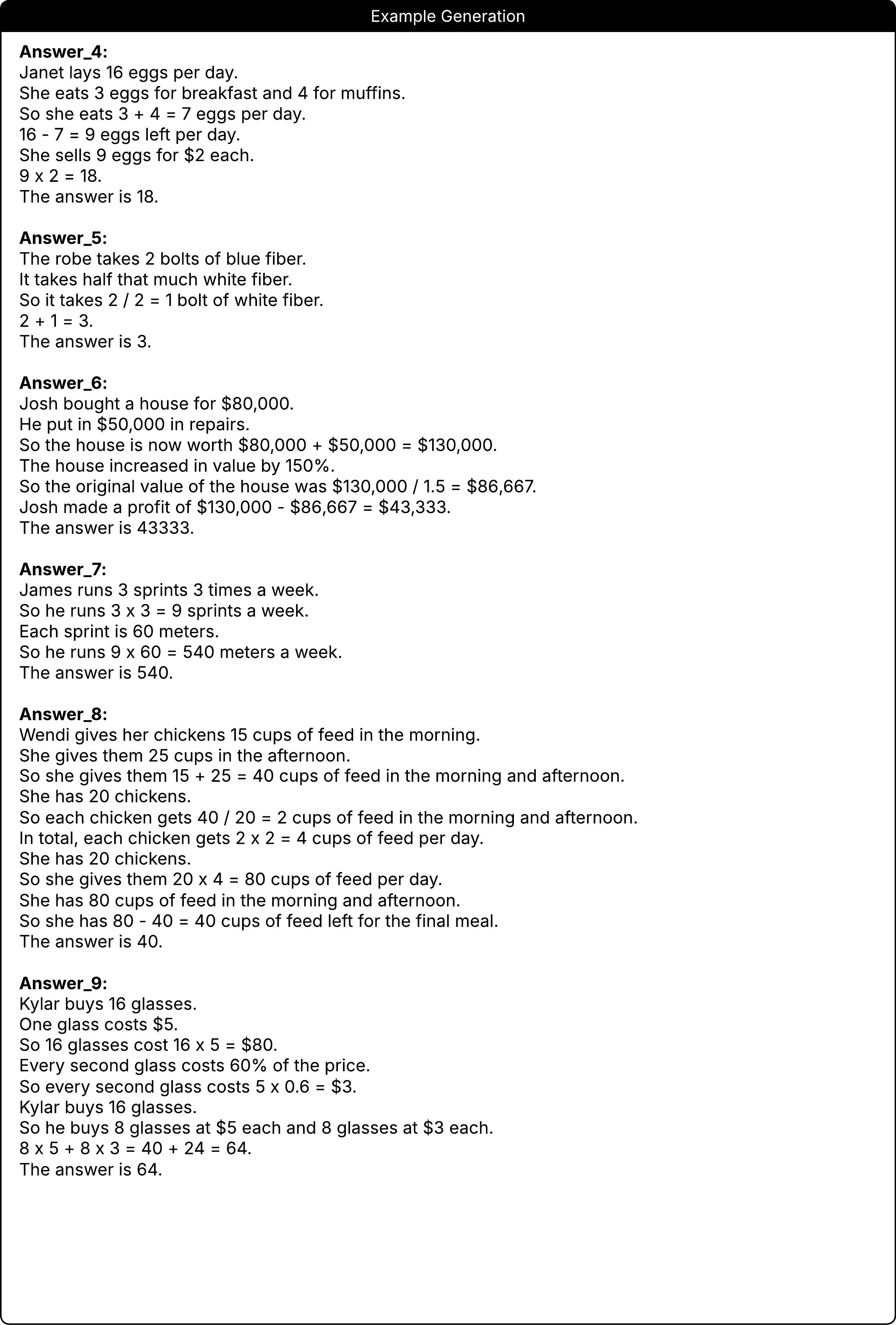}
    \label{fig:gsm8k_gen}
\end{figure}

\end{document}